\documentclass[12pt,linesnumbered,ruled,vlined]{article}
\usepackage{mathptmx}
\usepackage[latin9]{inputenc}
\usepackage{color}
\usepackage{mathrsfs}
\usepackage{mathtools}
\usepackage{algorithm2e}
\usepackage{amsmath}
\usepackage{amsthm}
\usepackage{amssymb}
\usepackage{graphicx}
\usepackage{appendix}
\usepackage[unicode=true,pdfusetitle,
 bookmarks=true,bookmarksnumbered=false,bookmarksopen=false,
 breaklinks=false,pdfborder={0 0 1},backref=false,colorlinks=true]
 {hyperref}
\hypersetup{
 pdfborderstyle=,pdfborderstyle={},pdfborderstyle={}}

\makeatletter
\@ifundefined{date}{}{\date{}}

\date{}

\usepackage{amsfonts}
\usepackage{dsfont}

\numberwithin{equation}{section}

\allowdisplaybreaks[4]
\DeclareMathAlphabet{\mathcal}{OMS}{cmsy}{m}{n}

\usepackage{xcolor}

\SetCommentSty{mycommfont}

\SetKwInput{KwInput}{Input}                
\SetKwInput{KwOutput}{Output}              
\SetKwInput{Hyperparameter}{Hyperparameters}
\SetKwInput{Parameter}{Parameters}

\makeatother

\begin{document}
\title{GAMMT: Generative Ambiguity Modeling Using Multiple Transformers}
\author{Xingcheng Xu\thanks{Shanghai AI Laboratory. Email: xingcheng.xu18@gmail.com}}
\maketitle
\begin{abstract}
{\normalsize{}
We introduce a novel model called GAMMT (Generative Ambiguity Models using Multiple Transformers) for sequential data that is based on sets of probabilities. Unlike conventional models, our approach acknowledges that the data generation process of a sequence is not deterministic, but rather ambiguous and influenced by a set of probabilities. To capture this ambiguity, GAMMT employs multiple parallel transformers that are linked by a selection mechanism, allowing for the approximation of ambiguous probabilities. The generative nature of our approach also enables multiple representations of input tokens and sequences. While our models have not yet undergone experimental validation, we believe that our model has great potential to achieve high quality and diversity in modeling sequences with uncertain data generation processes.
}{\normalsize\par}
\end{abstract}

\section{Introduction\label{sec:Introduction}}

Risk and ambiguity are two types of uncertainty that are commonly modeled. Risk refers to situations where the probabilities of various outcomes are known or can be estimated, while ambiguity refers to scenarios where the probability distribution of potential outcomes is unknown or uncertain.

In this paper, we present a novel approach for modeling data generation processes under ambiguity, which we call Generative Ambiguity Models using Multiple Transformers (GAMMT). Traditional machine learning methods assume data is sampled from a deterministic probability, but our proposed approach acknowledges that the process may be ambiguous and determined by multiple probabilities. Our GAMMT model utilizes multiple parallel transformers to model a set of probabilities in sequential data, such as token sequences in natural language. These parallel transformers function as a system to respond to ambiguity and provide multiple representations of the underlying input tokens and sequence.

To illustrate the concept of ambiguity, we use Ellsberg's urns, which were introduced in 1961 \cite{Ellsberg1961} and have stimulated research on ambiguity in economics and finance. Our GAMMT model leverages the effectiveness of transformers, which are attention-based mechanisms in deep neural networks, to model probabilities. Each parallel transformer receives an input sequence and outputs a deterministic probability, and the transformers are connected through a selection mechanism.

Our proposed GAMMT models have two key features that distinguish them from existing approaches. Firstly, they are generative and can effectively model ambiguity in data generation processes. Secondly, the last hidden layer of each parallel transformer provides multiple embeddings of each token and the input sequence, enabling diverse representations of the underlying input. We introduce the conceptual framework of our GAMMT models and anticipate that they will improve the quality and diversity of generated text, making it more engaging and human-like. Although we have not yet provided experimental validation, we believe that our models have great potential in this regard. Our approach to modeling ambiguity in sequential data is novel and provides a more interpretable and flexible method than existing techniques. The unique features of our model, such as the use of multiple transformers and its generative nature, make it a promising direction for future research in machine learning. We hope that this framework will stimulate further research in the fields of machine learning and natural language processing.

The paper is organized as follows: In Section \ref{sec:RelatedWork}, we review related work on transformers and ambiguity modeling. In Section \ref{sec:Framework}, we present our proposed notation for capturing the structure of uncertainty and define the set $\mathcal{P}^{\mathrm{LM}}$ to represent the set of plausible probability laws. We then introduce the GAMMT architecture in Section \ref{sec:Architecture}, which utilizes multiple parallel transformers linked by a selection mechanism to approximate ambiguous probabilities. In Section \ref{sec:Algorithms}, we provide an overview of the algorithms for the GAMMT model's architecture, model training, and inference. Finally, in Section \ref{sec:Conclusion}, we conclude this paper.

\section{Related Work\label{sec:RelatedWork}}

Transformers have emerged as a powerful and versatile tool in machine learning, driving significant progress in natural language processing (NLP), as demonstrated by models such as BERT (Devlin et al., 2018 \cite{DCL+2018}), OpenAI GPT (Radford et al., 2018 \cite{RNS+2018}), GPT-2 (Radford et al., 2019 \cite{RWC+2019}), GPT-3 (Brown et al., 2020 \cite{BMR+2020}), InstructGPT (Ouyang et al., 2022 \cite{OWJ+2022}), ChatGPT (\cite{OpenAI2022}), and Google T5 (Raffel et al., 2019 \cite{RSR+2019}), PaLM (Chowdhery et al., 2022 \cite{CND+2022}), etc. These models have achieved groundbreaking results and transformed NLP research. Additionally, transformers have extended their reach beyond NLP to other fields such as computer vision, exemplified by the success of Vision Transformers proposed by Dosovitskiy et al. in 2020 \cite{DBK+2020}, and even structural biology, as shown by AlphaFold 2 developed by Jumper et al. in 2021 \cite{JEP+2021}. This broad application of transformers highlights their versatility and potential for future breakthroughs.

Ambiguity is a widespread challenge in various domains, including vision, language, learning, and decision-making. In machine learning, several studies have explored the concept of ambiguity, which can arise from external sources or incomplete knowledge during the generation process.

Works by Ek et al. (2008) \cite{ERT+2008}, Patel et al. (2019) \cite{PN2019}, Yang et al. (2021) \cite{YZZ+2021}, Buisson et al. (2022) \cite{BAJB2022}, and others have previously explored ambiguity in machine learning. Ambiguity is pervasive in various fields, such as vision, natural language, general learning, and decision-making. For example, in images, visual relationships can be semantically ambiguous, as classified by Yang et al. \cite{YZZ+2021} into three types: synonymy ambiguity, hyponymy ambiguity, and multi-view ambiguity. In natural language, perceptual vagueness can occur at different levels of granularity, including words, sentences, paragraphs, and documents. Ambiguity in images or language can arise from external sources or due to individuals' feelings, perception, experiences, or incomplete knowledge during the generation process.

Traditionally, machine learning models have represented images, words, or sentences using a fixed feature vector in a deterministic way. However, such a fixed representation is insufficient to capture the ambiguity mentioned above. To address this limitation, researchers have proposed using multiple complementary representations of an underlying phenomenon. For example, Ek et al. \cite{ERT+2008} demonstrated the use of multi-modal regression on a benchmark human pose estimation dataset. Yang et al. \cite{YZZ+2021} utilized a probability distribution to represent each union region of an image using Gaussian embedding. In this approach, each union region is parameterized by a mean and variance, where the mean vector represents the normal feature vector, and the variance measures feature uncertainty.

While existing approaches, such as those proposed by Ek et al. \cite{ERT+2008} and Yang et al. \cite{YZZ+2021}, have made progress in handling ambiguity in machine learning, they do not explicitly model the probabilistic structure of ambiguity as in our proposed GAMMT model. Our model utilizes multiple parallel transformers to model probabilities in sequential data, providing a more interpretable and flexible approach to handling ambiguity. Additionally, the GAMMT model is generative, allowing it to effectively characterize ambiguity in the data generation process.

Compared to existing approaches, our proposed GAMMT model has several unique features that make it a promising avenue for future research in the field of machine learning. Firstly, it provides multiple ambiguous representations of the input tokens and sequence, allowing for greater flexibility in downstream tasks such as language understanding and generation. Secondly, the use of multiple transformers allows for a more comprehensive modeling of the complex and multi-dimensional nature of ambiguity in sequential data. Lastly, the generative nature of the GAMMT model provides additional insights into the generation process of ambiguous data.

Furthermore, in the field of neuroscience, cognitive neuroscientists have utilized neuroimaging techniques, such as functional magnetic resonance imaging (fMRI), to investigate brain activation and neuronal correlates in humans during decision-making under ambiguity. Studies conducted by Levy et al. (2010) \cite{LSN+2010}, Bach et al. (2011) \cite{BHP+2011}, Chumbley et al. (2012) \cite{CFB+2012}, Taya (2012) \cite{Taya2012}, and others have demonstrated that a distributed network of brain regions associated with cognitive control and reward processing is closely linked to decision-making in ambiguous situations. Notably, Chumbley et al. \cite{CFB+2012} observed clear ambiguity-dependent responses in the hippocampus, suggesting that certain neuronal systems may play a role in resolving ambiguity. 

This suggests that our GAMMT model, which is designed to model ambiguity in sequential data using multiple parallel transformers, may be able to simulate the distributed nature observed in cognitive neuroscience studies.

\section{Framework\label{sec:Framework}}

In the field of sequential modeling, the standard approach in machine learning involves factorizing joint probabilities $P(x)$ of a sequence $x=(s_{1},s_{2},\cdots,s_{\ell})\in V^{*}$, where $V^{*}=\cup_{\ell=0}^{\infty}V^{\ell}$ is a sequence space consisting of tokens from a vocabulary $V$. This factorization is commonly achieved by taking the product of conditional probabilities, as in popular language models such as GPT \cite{RNS+2018}, GPT-2 \cite{RWC+2019}, and GPT-3 \cite{BMR+2020}, etc. The joint probability can be expressed as:

\begin{equation}
P(x)=\prod_{n=1}^{\ell}P(s_{n}|s_{1},s_{2},\cdots,s_{n-1}),
\end{equation}
where $P(s_{n}|s_{1},s_{2},\cdots,s_{n-1}):=P(s_{1})$ when $n=1$.

For the conditional probability $P(x|z)$ of a pair of sequences $x=(s_{1},s_{2},\cdots,s_{\ell})\in V^{*}$ and $z\in V^{*}$, it can be factorized by taking the product of following conditionals, as in the original Transformer model \cite{VSP+2017}:

\begin{equation}
P(x|z)=\prod_{n=1}^{\ell}P(s_{n}|s_{1},s_{2},\cdots,s_{n-1},z),
\end{equation}
where $P(s_{n}|s_{1},s_{2},\cdots,s_{n-1},z):=P(s_{1}|z)$ when $n=1$.

In this section, we propose a novel approach to address sequential modeling problems. We recognize that ambiguity is prevalent in various scenarios, and the randomness of sequences may not be accurately described by a single probability distribution. Instead, we introduce a collection of probabilities to capture the structure of uncertainty associated with probabilities and develop a novel framework for sequential modeling that takes into account this collection of probabilities. Building on the work of Chen and Epstein (2022) \cite{CE2022}, we present notation that captures the structure of uncertainty and define the set $\mathcal{P}^{\mathrm{LM}}$ (see Section \ref{sec:ProbStructure}) to represent the set of plausible probability laws. Our approach recognizes that different sets of probabilities may be associated with different time points, allowing for a more complex modeling of sequential data.

\subsection{Uncertainty of Probabilities\label{sec:UncertainProb}}

In this subsection, we propose a novel approach to modeling uncertainty in sequential modeling problems by utilizing a collection of probabilities. This is analogous to having a mystical box containing multiple distinct dice, rather than just one. For each event or decision, we randomly select and throw one or more of these dice, and the resulting outcome is based on their combined outcomes. The data generation process is thus determined by a set of probabilities denoted by:
\begin{equation}
\mathcal{L}_n=\{P_{n1},P_{n2},\cdots,P_{nM}\},\quad n=1,2,\cdots
\end{equation}
where $\mathcal{L}_n$ represents the set of fundamental probability measures for creating the data process. The number of measures in the set determines the degree of model ambiguity or uncertainty. This novel approach enables us to model ambiguity in a more flexible and interpretable way than existing techniques.

The objective of sequence modeling is to maximize the product of conditional probabilities while also incorporating a selection mechanism, which is given by the following expression:

\begin{equation}
\prod_{n=1}^{\ell}S_{n}\left(\left\{ P_{nj}(s_{n}|s_{1},s_{2},\cdots,s_{n-1}),\ P_{nj}\in \mathcal{L}_n \right\}\right),
\end{equation}
where $P_{nj}(s_{n}|s_{1},s_{2},\cdots,s_{n-1}):=P_{nj}(s_{1})$ when $n=1$. The set $\mathcal{L}_n$ represents the collection of plausible probability laws for each one-step-ahead conditional at time point $n$. To model the collection of conditional probabilities $\mathcal{L}_n$, deep neural networks with learnable parameters $\theta_{n}$ are used, such as multiple Transformer decoders for natural language models in our paper. 

The selection mechanism $S_n$ is used to address the ambiguity structure present in the data. It may involve selecting probabilities at random subject to a uniform random variable or selecting the maximum of all possible conditionals. Notably, our proposed approach degenerates to the standard approach when $\mathcal{L}_n$ is a singleton set $\{P_n\}$.

Our GAMMT model captures the probabilistic structure of uncertainty by utilizing the set $\mathcal{P}^{\mathrm{LM}}$ (in the next subsection), which allows for different sets $\mathcal{L}_n$ of plausible probability laws for each one-step-ahead conditional at every time point. By incorporating this more flexible set of probability laws, the GAMMT model can capture a wider range of uncertainty and account for potential dependencies between experiments. In the next subsection, we will provide a detailed description of the probabilistic structure of uncertainty in our GAMMT model.


\subsection{Probabilistic Structure of Ambiguity\label{sec:ProbStructure}}

In this subsection, we present notation that captures the structure of uncertainty associated with probabilities, building on the work of Chen and Epstein (2022) \cite{CE2022}. When there is ambiguity about the probability law for each position in a sequence, understanding the relationship between positions becomes crucial, as this relationship can have significant implications for the interpretation of the results.

Consider a filtered space $(\Omega, \{\mathcal{G}_n\}_{n=1}^{\infty}))$, where $\Omega = \prod_{i=1}^{\infty}\Omega_i$ and each $\Omega_i$ represents the set of possible outcomes for the $i$th position in a sequence. Let $\Omega^{(n)}:=\prod_{i=1}^{n}\Omega_i$ and $\Omega_{(n+1)}:=\prod_{i=n+1}^{\infty}\Omega_i$. For each $n$, the $\sigma$-algebra $\mathcal{G}_n$ represents the observable tokens regarding positions $1, 2, \cdots, n$ on the underlying probability space $(\Omega^{(n)}, \mathcal{G}_n)$. We assume that $\mathcal{G}_n$ is increasing with $n$, with $\mathcal{G}_0$ being the trivial $\sigma$-algebra. The $\sigma$-algebra generated by all the observable tokens is denoted by $\mathcal{G}$ and is defined as the smallest $\sigma$-algebra that contains the sets $\mathcal{G}_n$, where $n$ ranges over the non-negative integers. In other words, $\mathcal{G}$ is the $\sigma$-algebra generated by the union of all the $\mathcal{G}_n$'s. The precise ex ante probabilities of observable tokens are unknown and are instead represented by a set $\mathcal{P}$ of probability measures on $(\Omega,\mathcal{G})$.

We provide the following additional notation and terminology:
\begin{equation*}
\omega^{(n)} = (\omega_1,\cdots,\omega_n)\in \Omega^{(n)},\ \omega_{(n+1)} = (\omega_{n+1},\omega_{n+2},\cdots)\in \Omega_{(n+1)},
\end{equation*}
\begin{equation*}
\mathcal{P}_{0,n} = \left\{P_{|\mathcal{G}_n}:\ P \in \mathcal{P}\right\},
\end{equation*}
\begin{equation*}
\mathcal{G}_{(n+1)} =\left \{A:\ A\subset \Omega_{(n+1)},\ \Omega^{(n)}\times A \in \mathcal{G}\right\}.
\end{equation*}
The set $\mathcal{P}_{0,n}$ represents the set of probability measures on the measurable space $(\Omega^{(n)},\mathcal{G}_n)$. The regular conditional probability of a measure $Q$ on the measurable space $(\Omega,\mathcal{G})$ given the sub-$\sigma$-algebra $\mathcal{G}_n$ is denoted by $Q(\cdot|\mathcal{G}_n)$. A probability kernel $\psi:\Omega^{(n)} \times \mathcal{G}_{(n+1)}\rightarrow[0,1]$ is called a $\mathcal{P}$\textit{-kernel} if, for every $n$ and every $\omega^{(n)}\in\Omega^{(n)}$, there exists a probability measure $Q\in \mathcal{P}$ such that 
\begin{equation}
\psi(\omega^{(n)},A)=Q(\Omega^{(n)}\times A|\mathcal{G}_n)(\omega^{(n)})
\end{equation}
for all $A\in\mathcal{G}_{(n+1)}$. 

A probability measure $P$ on $(\Omega,\mathcal{G})$ is said to be \textit{rectangular} with respect to the filtration ${\mathcal{G}_n}$ if, for every $n$, every $P_n\in \mathcal{P}_{0,n}$ and every $\mathcal{P}$-kernel $\psi$ as above, the measure $P$ defined by
\begin{equation}
P(A)=\int_{\Omega^{(n)}}\int_{\Omega_{(n+1)}}I_{A}\left(\omega^{(n)},\omega_{(n+1)}\right)\psi\left(\omega^{(n)},d\omega_{(n+1)}\right)P_n\left(d\omega^{(n)}\right),\quad \forall A\in\mathcal{G},
\end{equation}
belongs to $\mathcal{P}$.

If the probability set $\mathcal{P}$ consists of a single probability distribution $P$, then the principle of rectangularity is a straightforward consequence of the Bayesian updating rule. This is because $P$ can be decomposed into its marginal and conditional distributions, which can then be recombined to obtain $P$ again. However, when dealing with more complex sets $\mathcal{P}$, rectangularity is a non-trivial requirement. It demands that $\mathcal{P}$ is closed under the operation of pasting together conditionals and marginals that may have been induced by different measures within $\mathcal{P}$. The significance of rectangularity is highlighted in the reference \cite{CE2022}.

Specialize the above framework by assuming that there exists a measurable space $(\overline{\Omega}, \overline{\mathcal{F}})$ such that, for all $1 \leq i \leq n$, $(\Omega_i, \mathcal{F}_i) = (\overline{\Omega}, \overline{\mathcal{F}})$ and $\mathcal{G}_n = \prod_{i=1}^n \mathcal{F}_i$. That is, all positions in a sequence have a common set of possible outcomes $\overline{\Omega}$ and an associated common $\sigma$-algebra $\overline{\mathcal{F}}$. As an example, for a language modeling task, $\overline{\Omega}$ could be the set $V$ of all possible tokens in a given language, i.e. $\overline{\Omega}=V$, and $\overline{\mathcal{F}}$ could be the collection of all subsets of $\overline{\Omega}$ that can be generated by using the given grammar rules and vocabulary $V$ of the language.

We can define the \textit{one-step-ahead conditionals} as follows: for each $P \in \mathcal{P}$ and each $n$, we let $P_{n,n+1}(\omega^{(n)})$ be the restriction to $\mathcal{G}_{n+1}$ of $P\left(\cdot|\mathcal{G}_n\right)(\omega^{(n)})$. Assuming that there exists a subset $\mathcal{L}$ of $\Delta(\overline{\Omega}, \overline{\mathcal{F}})$, all of whose measures are equivalent, we can specialize the above framework by defining the indistinguishably and independently distributed (IID) model through the set $\mathcal{P}^{\mathrm{IID}}$. Specifically,
\begin{equation}
\mathcal{P}^{\mathrm{IID}} = \left\{ P \in \Delta(\Omega, \mathcal{G}):\ P_{n,n+1}(\omega^{(n)}) \in \mathcal{L},\ \forall n,\ \forall \omega^{(n)} \in \Omega^{(n)} \right\}.
\end{equation}
The set $\mathcal{P}^{\mathrm{IID}}$ is comprised of measures for which the one-step-ahead conditionals, at every history, are in $\mathcal{L}$. In other words, $\mathcal{L}$ represents the plausible probability laws for every position in a seqeunce, reflecting our partial ignorance about each position individually, regardless of its history.

The set $\mathcal{P}^{\mathrm{IID}}$ only requires that each one-step-ahead conditional belongs to the set of plausible probability laws $\mathcal{L}$, without imposing any further constraints on the measures of the entire sequence. It models partial ignorance about each position separately without considering the history. 

However, it is possible to have a more complex situation where the one-step-ahead conditionals at each time point/position have different sets of probabilities, denoted by $\mathcal{L}_n$. In this case, the set $\mathcal{P}^{\mathrm{LM}}$ is defined similarly to $\mathcal{P}^{\mathrm{IID}}$, but with the requirement that each one-step-ahead conditional belongs to the corresponding set of plausible probability laws $\mathcal{L}_n$. That is, 
\begin{equation}
\mathcal{P}^{\mathrm{LM}} = \left\{ P \in \Delta(\Omega, \mathcal{G}):\ P_{n,n+1}(\omega^{(n)}) \in \mathcal{L}_n,\ \forall n,\ \forall \omega^{(n)} \in \Omega^{(n)} \right\}.
\end{equation}

Our GAMMT model captures the probabilistic structure of uncertainty by utilizing the set $\mathcal{P}^{\mathrm{LM}}$. Specifically, the GAMMT model allows for the possibility of different sets of plausible probability laws for each one-step-ahead conditional at every time point, as denoted by $\mathcal{L}_n$. By incorporating this more flexible set of probability laws, the GAMMT model is able to capture a wider range of uncertainty and account for potential dependencies between different positions in a sequence. 


\section{Architecture\label{sec:Architecture}}

In this section, we propose a class of deep neural networks that capture ambiguity in the data-generating process. Specifically, we introduce the GAMMT model, which utilizes multiple self-attention Transformer decoders commonly used in natural language processing tasks (e.g., \cite{BMR+2020, DCL+2018, PH2022, RNS+2018, RWC+2019, RSR+2019, VSP+2017}).

\begin{center}
\begin{figure}[!htbp]
\begin{centering}
\includegraphics[width=13cm]{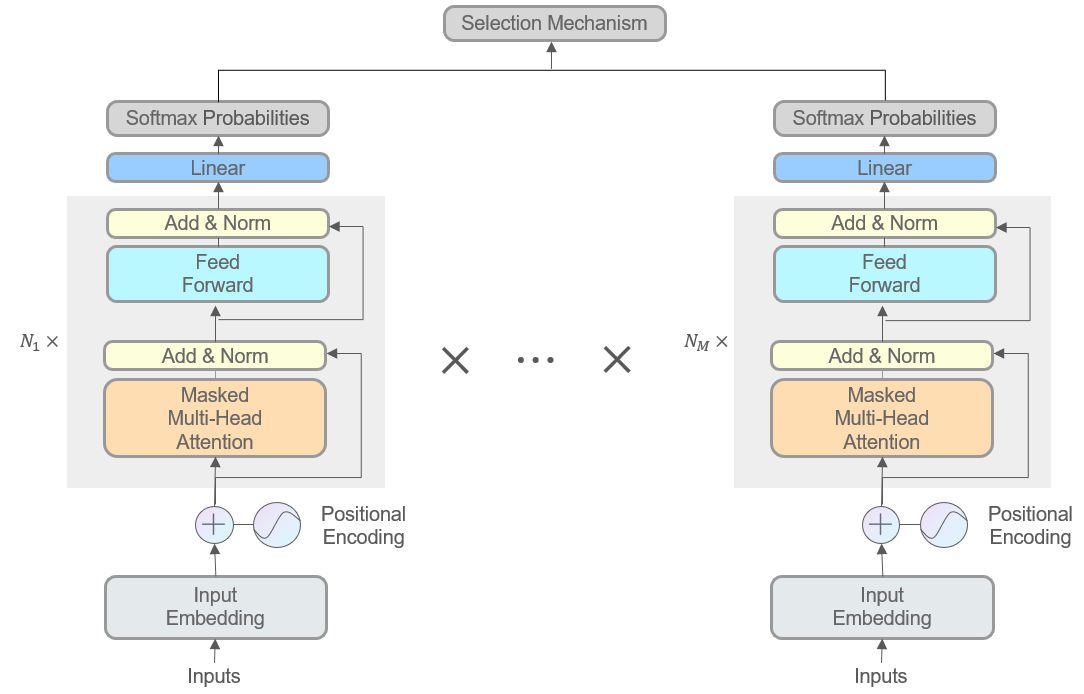}
\par\end{centering}
\caption{Model architecture - The Transformer decoders to approximate sets
of probabilities}
\label{fig_architecture}
\end{figure}
\end{center}

The architecture of the GAMMT model, shown in Figure \ref{fig_architecture}, consists of $M$ parallel Transformer decoders, each with $N$ identical layers. Each layer includes a masked multi-head self-attention sublayer with $h$ parallel attention heads, a feed-forward network, and layer normalization. The output of each decoder is transformed into a set of next-token probabilities, denoted by $\mathcal{L}_n$ for the $n$-th position in a sequence, using a linear transformation and softmax function. The $M$ Transformer decoders are connected by a selection mechanism $S_n$ that captures the ambiguity in the data-generating process.

By utilizing the set $\mathcal{P}^{\mathrm{LM}}$ of plausible probability laws, the GAMMT model captures a wide range of uncertainty and potential dependencies between different positions in a sequence. The precise probabilistic structure of the GAMMT model is described in Section \ref{sec:ProbStructure}. The model learns the conditional probabilities at each time point $n$, which are then combined using the selection mechanism $S_n$ to capture the ambiguity in the data-generating process.


\vspace*{\fill}\vspace*{\fill}

\newpage{}

\section{Algorithms\label{sec:Algorithms}}

This section provides an overview of the algorithms for the proposed GAMMT model's architecture, model training, and inference. These algorithms are presented to help readers understand the workings of the GAMMT model. The pseudocode for the algorithms described in this section can be found in \ref{appendix_a}. 

Algorithm \ref{alg:architecture} outlines the forward pass of the model, which generates a set of probability matrices for predicting the next token, given an input sequence of tokens $x$ with length $\ell_{x}$. The vocabulary of tokens is denoted by $V$, with $N_{V}$ being the size of the vocabulary. The operations of embedding and Transformer decoding are similar to those described in Algorithms 1, 2, and 10 of Phuong and Hutter's work \cite{PH2022}. 
Algorithm \ref{alg:training} presents the training process for the GAMMT model. The $M$ parallel Transformer decoders are trained simultaneously using the given selection mechanism for $N_{epochs}$ epochs to obtain the trained parameters. 
Algorithm \ref{alg:inference} outlines the process of generating model predictions for a prompt using the GAMMT model, with the specific sampling method dependent on the selection mechanism used during training.

For brevity, we omit the details of the initial token and positional embedding, as well as the specifics of the Transformer decoders. Interested readers can refer to Phuong and Hutter's work \cite{PH2022} for a comprehensive overview of Transformer models and related approaches.

%
%

\section{Conclusion\label{sec:Conclusion}}

In conclusion, this paper presents a novel approach for modeling data generation processes under ambiguity, called GAMMT. Our proposed approach acknowledges that the process may be ambiguous and determined by multiple probabilities, which is different from traditional machine learning methods that assume data is sampled from a deterministic probability. GAMMT utilizes multiple parallel transformers to model a set of probabilities in sequential data, providing multiple representations of the underlying input tokens and sequence.

By leveraging the effectiveness of transformers, our GAMMT model has two key features that distinguish it from existing approaches. Firstly, it is generative and can effectively model ambiguity in data generation processes. Secondly, the last hidden layer of each parallel transformer provides multiple embeddings of each token and the input sequence, enabling diverse representations of the underlying input. We anticipate that our models will improve the quality and diversity of generated text, making it more engaging and human-like. Although we have not yet provided experimental validation, we believe that our models have great potential in this regard.

Our approach to modeling ambiguity in sequential data is novel and provides a more interpretable and flexible method than existing techniques. The unique features of our model, such as the use of multiple transformers and its generative nature, make it a promising direction for future research in machine learning. We hope that this framework will stimulate further research in the fields of machine learning and natural language processing, and that our work will contribute to the development of more effective and flexible models for handling ambiguity in data generation processes.

%

\newpage{}

\begin{appendices}

\renewcommand{\thesection}{Appendix A}
\section{Algorithms}\label{appendix_a}

\renewcommand{\thesection}{A}



\vspace*{\fill}
This appendix outlines the algorithms used for the proposed method, including model architecture, training, and inference. The GAMMT model architecture, which details the forward pass of the model, is presented in Algorithm \ref{alg:architecture}. The model takes an input sequence of tokens $x$ with a length of $\ell_{x}$, and outputs a set of probability matrices for predicting the next token. The vocabulary of tokens is denoted by $V$, with $N_{V}$ representing the vocabulary size. The embedding and Transformer decoding operations are similar to those described in Phuong and Hutter's work \cite{PH2022}, as outlined in Algorithms 1, 2, and 10 (\cite{PH2022}).
\begin{center}
\begin{algorithm}[!h]
\DontPrintSemicolon \SetAlgoLined

\caption{$\mathcal{P}\leftarrow\textrm{DTransformers}(x|\theta)$}
\label{alg:architecture}

\tcc{Model architecture - the Transformer decoders, forward pass}

\tcc{Input: a sequence of token IDs with length $\ell_{x}$} \tcc{Output:
probabilities of next token} \KwInput{$x\in V^{*}$} \KwOutput{$P_{\theta_{m}}\in[0,1]^{N_{V}\times\ell_{x}},$
$m\in[M]:=\{1,2,\cdots,M\}$ and the Selection $S$}

\Hyperparameter{The number of Transformer decoders $M$, the Transformer
decoders Layers $N_{j}$, heads $H_{j}$, input embedding dimension
$d_{e}$, hidden layer dimension $d_{mlp}$, maximum length of input
$\ell_{max}$, selection $S$} \Parameter{$\theta=({\theta_{m}})_{m\in[M]}$
includes all token and positional embedding matrices, all Transformers'
parameters}

\For{$m=1,2,\cdots,M$}{ $X_{m}\leftarrow$ TransformerDecoder($m$,Embedding($x$))\;
$P_{\theta_{m}}=\textrm{softmax}(W_{o}^{(m)}X_{m})$\; }

\Return{$(P_{\theta_{m}})_{m\in[M]}$, $S=S(P_{\theta_{1}},\cdots,P_{\theta_{M}})$}
\end{algorithm}
\end{center}
\vspace*{\fill}

\newpage

\vspace*{\fill}
The training process for the GAMMT model is presented in Algorithm \ref{alg:training}. The model employs $M$ parallel Transformer decoders, trained simultaneously with a selection mechanism. The trained parameters are obtained after $N_{epochs}$ epochs of training.
\begin{center}
\begin{algorithm}[!h]
\DontPrintSemicolon \SetAlgoLined

\caption{$\hat{\theta}\leftarrow\textrm{Training}(x_{1:N_{data}},\theta)$}
\label{alg:training}

\tcc{Model training - prediction of the next token}

\tcc{Input: a dataset of sequences and initial parameters} \tcc{Output:
the trained parameters} \KwInput{$\{x_{n}\}_{n=1}^{N_{data}}$,
$\theta=({\theta_{m}})_{m\in[M]}$} \KwOutput{$\hat{\theta}=({\hat{\theta}_{m}})_{m\in[M]}$}

\Hyperparameter{Epochs $N_{epochs}$, learning rate $\eta>0$ and
the selection mechanism $S$}

\For{$i=1,2,\cdots,N_{epochs}$}{ \For{$n=1,2,\cdots,N_{data}$}{
$\ell_{x}\leftarrow\textrm{length}(x_{n})$\; $(P_{\theta_{m}})_{m\in[M]},\_\leftarrow\textrm{DTransformers}(x_{n}|\theta)$\;
$\textrm{loss}(\theta)=-\sum_{t=1}^{\ell_{x}-1}\log S\left(\left\{ P_{\theta_{m}}[x_{n}[t+1],t]\right\} _{m=1}^{M}\right)$\;
$\theta\leftarrow\theta-\eta\cdot\nabla\textrm{loss}(\theta)$\;
} }

\Return{$\hat{\theta}=\theta$}
\end{algorithm}
\end{center}
\vspace*{\fill}

\newpage

\vspace*{\fill}
Algorithm \ref{alg:inference} outlines the process of generating model predictions from a given prompt using the GAMMT model. The output is a continuation of the prompt generated by the model, with the sampling method used during inference depending on the selection mechanism employed during training.
\begin{center}
\begin{algorithm}[!h]
\DontPrintSemicolon \SetAlgoLined

\caption{$y\leftarrow\textrm{Inference}(x,\hat{\theta})$}
\label{alg:inference}

\tcc{Model inference - generate a sequence based on the trained
model and a prompt}

\tcc{Input: the trained parameters and a prompt} \tcc{Output:
a continuation of the prompt sampled from the trained model} \KwInput{$\hat{\theta}=({\hat{\theta}_{m}})_{m\in[M]}$,
$x\in V^{*}$} \KwOutput{$y\in V^{*}$} \Hyperparameter{temprature
$\tau>0$}

$\ell_{x}\leftarrow\textrm{length}(x)$\; $y\leftarrow\emptyset$\;
\While{$y\neq\textrm{eos\_token}$}{ $(P_{\theta_{m}})_{m\in[M]},\_\leftarrow\textrm{DTransformers}(x|\hat{\theta})$\;
\If{$S\sim\mathcal{R}([M])$}{ $u\leftarrow$ sample a random
variable $\mathcal{R}$ on $[M]$\; $p\leftarrow P_{\hat{\theta}_{u}}[:,\ell_{x}]$\;
sample a token $y$ from the probability $q\propto p^{1/\tau}$\;
$x\leftarrow[x,y]$\; $\ell_{x}\leftarrow\ell_{x}+1$\; } \ElseIf{$S=\max$}{
\For{$m=1,2,\cdots,M$}{ $p_{m}\leftarrow P_{\hat{\theta}_{m}}[:,\ell_{x}]$\;
sample a token $y_{m}$ from the probability $q_{m}\propto p_{m}^{1/\tau}$\;
} $y\leftarrow y_{m^{*}}$ with $m^{*}=\textrm{argmax}\left\{ q_{m}(y_{m}),\ m\in[M]\right\} $
\; $x\leftarrow[x,y]$\; $\ell_{x}\leftarrow\ell_{x}+1$\; } 
 } \Return{$y=x$} 
\end{algorithm}
\end{center}
\vspace*{\fill}



\end{appendices}


\begin{thebibliography}{10}

\bibitem{Ama2023} Amatriain X. Transformer models: an introduction and catalog. arXiv preprint arXiv:2302.07730, 2023.

\bibitem{BHP+2011} Bach D R, Hulme O, Penny W D, et al. The known
unknowns: neural representation of second-order uncertainty, and ambiguity.
Journal of Neuroscience, 2011, 31(13): 4811-4820.

\bibitem{BMR+2020}Brown T, Mann B, Ryder N, et al. Language models
are few-shot learners. NeurIPS, 2020, 33: 1877-1901.

\bibitem{BAJB2022} Buisson M, Alonso-Jim\'{e}nez P, Bogdanov D. Ambiguity
Modelling with Label Distribution Learning for Music Classification.
IEEE International Conference on Acoustics, Speech and Signal Processing
(ICASSP), 2022: 611-615.

\bibitem{CW1992} Camerer C, Weber M. Recent developments in modeling
preferences: Uncertainty and ambiguity. Journal of risk and uncertainty,
1992, 5(4): 325-370.

\bibitem{CE2022} Chen Z, Epstein L G. A central limit theorem for sets of probability measures. Stochastic Processes and their Applications, 2022, 152: 424-451.

\bibitem{CTJ+2021} Chen M, Tworek J, Jun H, et al. Evaluating large language models trained on code. arXiv preprint arXiv:2107.03374, 2021.

\bibitem{CND+2022}  Chowdhery A, Narang S, Devlin J, et al. PaLM: Scaling language modeling with pathways. arXiv preprint arXiv:2204.02311, 2022.

\bibitem{CFB+2012} Chumbley J R, Flandin G, Bach D R, et al. Learning
and generalization under ambiguity: an fMRI study. PLoS Computational
Biology, 2012, 8(1): e1002346.

\bibitem{DCL+2018} Devlin J, Chang M W, Lee K, et al. BERT: Pre-training
of deep bidirectional transformers for language understanding. arXiv:1810.04805,
2018. NAACL, 2019.

\bibitem{DBK+2020} Dosovitskiy A, Beyer L, Kolesnikov A, et al. An
Image is Worth 16x16 Words: Transformers for Image Recognition at
Scale. arXiv:2010.11929, 2020. 

\bibitem{ERT+2008} Ek C H, Rihan J, Torr P H S, et al. Ambiguity
modeling in latent spaces. International workshop on Machine Learning
for Multimodal Interaction. Springer, Berlin, Heidelberg, 2008: 62-73.

\bibitem{Ellsberg1961} Ellsberg, D. Risk, Ambiguity, and the Savage
Axioms, Quarterly Journal of Economics, 1961, 75, 643-669.

\bibitem{FZS2022} Fedus W, Zoph B, Shazeer N. Switch transformers: Scaling to trillion parameter models with simple and efficient sparsity. The Journal of Machine Learning Research, 2022, 23(1): 5232-5270.

\bibitem{HBM+2022} Hoffmann J, Borgeaud S, Mensch A, et al. Training compute-optimal large language models. arXiv preprint arXiv:2203.15556, 2022.

\bibitem{IS2022} Ilut C L, Schneider M. Modeling uncertainty as ambiguity:
A review. NBER Working Paper, 2022.

\bibitem{JEP+2021} Jumper J, Evans R, Pritzel A, et al. Highly accurate
protein structure prediction with AlphaFold. Nature 596, 583-589,
2021.

\bibitem{LSN+2010} Levy I, Snell J, Nelson A J, et al. Neural representation
of subjective value under risk and ambiguity. Journal of neurophysiology,
2010, 103(2): 1036-1047.

\bibitem{OWJ+2022} Ouyang L, Wu J, Jiang X, et al. Training language models to follow instructions with human feedback[J]. Advances in Neural Information Processing Systems, 2022, 35: 27730-27744.

\bibitem{OpenAI2022}  OpenAI (2022), ChatGPT: https://openai.com/blog/chatgpt

\bibitem{PN2019} Patel R, Nenkova A. Modeling ambiguity in text:
A corpus of legal literature. 2019.

\bibitem{PH2022} Phuong M, Hutter M. Formal algorithms for transformers.
arXiv:2207.09238, 2022. 

\bibitem{RNS+2018} Radford A, Narasimhan K, Salimans T, et al. Improving
language understanding by generative pre-training. 2018.

\bibitem{RWC+2019} Radford A, Wu J, Child R, et al. Language models
are unsupervised multitask learners. OpenAI, 2019.

\bibitem{RBC+2021} Rae J W, Borgeaud S, Cai T, et al. Scaling language models: Methods, analysis \& insights from training gopher. arXiv preprint arXiv:2112.11446, 2021.

\bibitem{RSR+2019} Raffel C, Shazeer N, Roberts A, et al. Exploring
the Limits of Transfer Learning with a Unified Text-to-Text Transformer.
arXiv:1910.10683, 2019.

\bibitem{SFA+2022} Scao T L, Fan A, Akiki C, et al. BLOOM: A 176B-parameter open-access multilingual language model. arXiv preprint arXiv:2211.05100, 2022.

\bibitem{Taya2012} Taya F. Seeking ambiguity: a review on neuroimaing
studies on decision making under ambiguity. Recherches \'Economiques
de Louvain/Louvain Economic Review, 2012, 78(03-04): 85-100.

\bibitem{TKC+2022} Taylor R, Kardas M, Cucurull G, et al. Galactica: A large language model for science. arXiv preprint arXiv:2211.09085, 2022.

\bibitem{TDH+2022} Thoppilan R, De Freitas D, Hall J, et al. LaMDA: Language models for dialog applications. arXiv preprint arXiv:2201.08239, 2022.

\bibitem{TLI+2023} Touvron H, Lavril T, Izacard G, et al. LLaMA: Open and efficient foundation language models. arXiv preprint arXiv:2302.13971, 2023.

\bibitem{VSP+2017} Vaswani A, Shazeer N, Parmar N, et al. Attention
is all you need. NeurIPS, 2017, 30.

\bibitem{WBZ+2021} Wei J, Bosma M, Zhao V Y, et al. Finetuned language models are zero-shot learners. arXiv preprint arXiv:2109.01652, 2021.

\bibitem{YDY+2019} Yang Z, Dai Z, Yang Y, et al. Xlnet: Generalized autoregressive pretraining for language understanding. Advances in neural information processing systems, 2019, 32.

\bibitem{YZZ+2021} Yang G, Zhang J, Zhang Y, et al. Probabilistic
modeling of semantic ambiguity for scene graph generation. Proceedings
of the IEEE/CVF Conference on Computer Vision and Pattern Recognition.
2021: 12527-12536.

\bibitem{ZLD+2022} Zeng A, Liu X, Du Z, et al. GLM-130B: An open bilingual pre-trained model. arXiv preprint arXiv:2210.02414, 2022.

\bibitem{ZRG+2022} Zhang S, Roller S, Goyal N, et al. OPT: Open pre-trained transformer language models. arXiv preprint arXiv:2205.01068, 2022.

\end{thebibliography}
\end{document}